\documentclass[letterpaper, 10 pt, conference]{ieeeconf}  

\IEEEoverridecommandlockouts                              

\usepackage[utf8]{inputenc} 
\usepackage[T1]{fontenc}    
\usepackage{hyperref}       
\usepackage{url}            
\usepackage{booktabs}       
\usepackage{amsfonts}       
\usepackage{nicefrac}       
\usepackage{microtype}      
\usepackage{graphicx}
\usepackage{amsmath}

\title{Zero Shot Learning on Simulated Robots}

\begin{document}

%
\author{Robert Kwiatkowski$^{1}$ and Hod Lipson$^{2}$\\
Columbia University\\
\tt\small \{robert.kwiatkowski, hod.lipson\}@columbia.edu
\thanks{$^{1}$Department of Computer Science}%
\thanks{$^{2}$Department of Mechanical Engineering}%
}
\maketitle
\thispagestyle{empty}
\pagestyle{empty}

\begin{abstract}
In this work we present a method for leveraging data from one source to learn how to do multiple new tasks. Task transfer is achieved using a self-model that encapsulates the dynamics of a system and serves as an environment for reinforcement learning. To study this approach, we train a self-models on various robot morphologies, using randomly sampled actions. Using a self-model, an initial state and corresponding actions, we can predict the next state. This predictive self-model is then used by a standard reinforcement learning algorithm to accomplish tasks without ever seeing a state from the "real" environment. These trained policies allow the robots to successfully achieve their goals in the "real" environment. We demonstrate that not only is training on the self-model far more data efficient than learning even a single task, but also that it allows for learning new tasks without necessitating any additional data collection, essentially allowing zero-shot learning of new tasks.
\end{abstract}

\section{Introduction}
The field of robotic control has made massive strides in recent years through leveraging Deep Learning and data driven approaches to solve problems that have plagued roboticists for years. However, Deep Learning in robotics suffers from a big problem: Deep Learning needs large amounts of data and data is difficult to generate in the real world. Nearly all modern solutions rely on hand coded simulators like those proposed by \cite{gazebo} which, while often very good are labor intensive to construct, cannot adapt to changing circumstances, and still suffer from the "sim-to-real" gap.

In this work, we propose a data driven method to solve these problems. We propose a method of learning a "self-model" a model which learns the dynamics of a robotic system to predict future sensor outputs, which can accurately and precisely predict a sequence of future states given a starting state and a sequence of actions. In order to learn this self-model, we can use any state-action-state tuple meaning that unlike in other modern data driven control methods, every byte of data collected can be used to improve the model. We can then use this self-model to produce an infinite amount of artificial data to feed Deep Learning models as they learn a wide variety of desired tasks.

If a self-model is sufficiently predictive, a planner or reinforcement learning agent would essentially be able to look into the future by inputting actions and receiving future states. Such a sufficiently predictive self-model would allow for multiple agents trained on a variety of tasks without the need for any additional data beyond what is required to train the self-model. This would give the appearance of zero-shot learning as, from the outside the robot would gather data to train its self-model, train the self-model, learn a policy, and then be able to execute that policy successfully with the only bottleneck being computation. 

\begin{figure}[h]
	\begin{center}
	\includegraphics[scale=0.5]{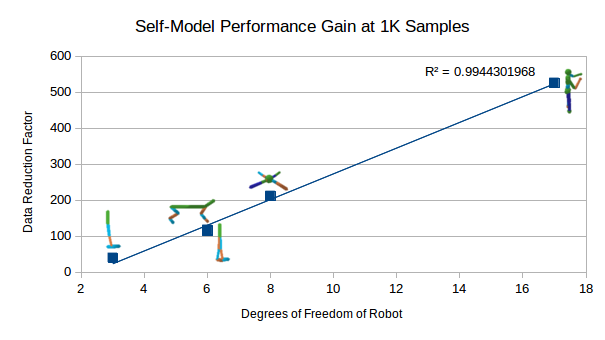}
	\end{center}
	\caption{Using self-model reduces training data requirements by two to three orders of magnitude. The more complex the robot the higher the gain factor}
	\label{eff_chart}
\end{figure}

\begin{figure*}[h]
\hspace*{0.3in}
	\begin{center}
	\includegraphics[scale=0.45]{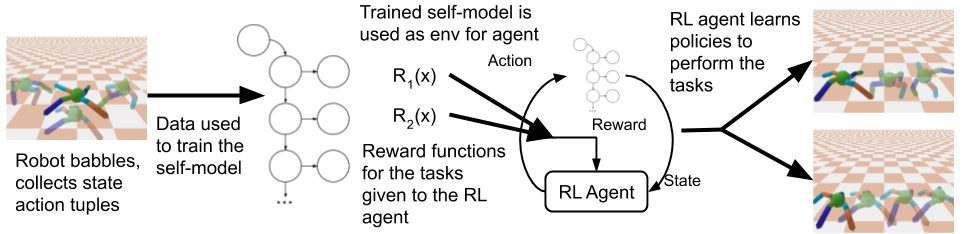}
	\end{center}
	\caption{Overview of the self-modeling learning process. A robot of unknown morphology moves randomly collecting data about itself. This data is then used to train a self-model which learns the dynamics of the system and is able to predict the next states in open loop. This self-model is then used in conjunction with a reward signal to train a reinforcement learner to accomplish an arbitrary task.}
	\label{ant}
\end{figure*}
\section{Related Work}

\subsection{Learning Robot Dynamics}
Work has been done in the past on learning robot dynamics and leveraging those models to make robots accomplish their tasks. \cite{josh_science} worked on a quadrapedal walking robot, however notably a quadraped with a less intuitive motor layout. This work created an algorithm for generating a physical model of the quadraped robot. The model they created was sufficient to generate a walking gait. However, this paper crucially included a simulator with which to generate the physical model. This simulator required a knowledge of the physics and mechanics of the world and so is not a fully data driven approach. \cite{rob_science} also worked to generate a self-model of a robot that was sufficient to do planning on. While the model was accurate enough to do a variety of tasks the robot used was far less complex and no task was learned. \cite{preco} too used machine learning to model the dynamics of a robotic system. For this paper the authors chose a robotic hand, a platform with significant complexity. However, this paper did not test their models as rigorously as the other works as they did not use their model for a variety of complex tasks. Ultimately, to our knowledge there has not been any work that has created a self-model of a robot with sufficient precision and accuracy to allow for not only task planners, but for learning agents to learn policies for acomplishing their tasks completely on the self-model. 

\subsection{Limited Action Space Environment Learning for Reinforcement Learning}
Learning environments in order to improve reinforcement learning performance is becomming popular due to their great successes. \cite{world_models} successfully leverage the use of models of reinforcement learning environments to train agents. This work however fully integrated their world models with the agents so that the agent would make use of the latent space of the world model as opposed to some predicted state $\hat{S}$. \cite{recurrent_env_sims} also used recurrent models to learn a model of their environment. This work does an extensive reviews of a number of different ways to effectively predict the environment as well as work on how to integrate it with reinforcement learning in the discrete action space. The authors touch on some elements that translate well to using environment learning to create self-models such as the learning of prediction independent simulators. \cite{ref_i2a} also proposes an architecture for leveraging environment learning to make a more efficient reinforcement learning algorithm. This paper showed success in the Sokoban environment however still had a limited dimensional action space and as such much simpler world dynamics. \cite{policy_opt_exp} similarly explored methods for merging model based and model free reinforcement learning together in to solve discrete action space reinforcement learning problems. \cite{model_based_atari} has proposed a method similar to the one we present in this paper. \cite{model_based_atari} gathered data about the world and then used a world model to train a reinforcement learner. However this paper was still fairly limited learning a specific reward function as well as using environments with a very limited discrete action space.

A critical difference between our approach and other environment learning approaches mentioned here however is the difference in domain. The aforementioned environment learning work has been done in the visual domain, taking in images and often outputting predicted image frames and works with relatively small and discrete action spaces. Our work however is in the domain of robot sensor readings and as such deals exclusively with continuous real value states and actions. As such the observable state and action space becomes significantly larger. 

\subsection{Environment Learning in the Continuous Domain}
There has been, however, some work on learning environment models for the continuous domain. \cite{trpo} presents another hybrid of model based and model free reinforcement learning to learn an environment model and leverage that along with the reinforcemnt learner to improve them both in step while learning to accomplish a task. A crucial difference is that this work combines the learned models with the reinforcement learning algorithm creating a hybrid approach. This approach is usually not as transferable to any arbitrary reinforcement learning algorithm and leaves them inflexible to solving any problem that they have not already been trained on. This inflexibility leads to these approaches essentially "throwing away" all the data they have previously seen when presented with a new task as they will have to learn a new policy from scratch. \cite{uncert_imagination} however does use artificial data to augment the real data that a reinforcement learner gets while training. While this paper is similar to our methods they use relatively simple robots focusing on robotic arms and do not fully train their reinforcement learner on artificial data limiting its zero-shot potential. 

\section{Experimental Design}
For all of the experiments in this paper we used a simulated robot to collect data through random motion. This data was later used to train a self-model on which the task planner learned to execute a task without any additional observations from the simulator. All experiments were conducted on an Nvidia 1080 TI GPU.
\subsection{PyBullet Environments} 


\begin{figure}
	\hspace*{0.2in}
	\begin{center}
	\begin{tabular}{ccc}
		\includegraphics[scale=0.45]{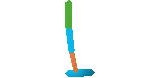}&
		\includegraphics[scale=0.45]{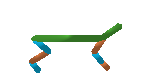}&
		\includegraphics[scale=0.45]{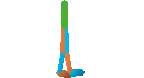}\\
		Hopper & HalfCheetah & Walker2D\\
		\includegraphics[scale=0.3]{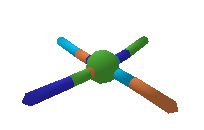}&&
		\includegraphics[scale=0.45]{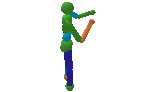}\\
		Ant &  & Humanoid\\
	\end{tabular}
	\end{center}
	\caption{Pictures of each of the different robot morphologies we tested in our experiments.}
	\label{robots}
\end{figure}
For this work, we collected all data and ran all experiments on the PyBullet \cite{pybullet} simulation's varied robot morphologies. These robots have been used in simulation in the past and have proven themselves to be an interesting platform to work with. For example The Ant has shown itself to be sufficiently versatile that it has been able to run \cite{MAML}, as well as jump and navigate obstacles \cite{ant_obstacles}, while the Humanoid has shown itself to be a relatively difficuly control task. 

For all of these robot morphologies, the actions exist in the range $[-5, 5]$ and so in our experiments we normalize them to fall within $[-1, 1]$. The state space however is more complex. Each state contain $2m+f+8$ continuous values, where $m$ corresponds to the number of motors and $f$ corresponds to the number of feet, This state space corresponds to the change in $z$ position, the sine and cosine of the angle of the robot to the predefined "target" position, the velocities in the x, y, and z direction, the roll and pitch, the speed and position of each of the joints, as well as information regarding the feet contacts. 

For the purposes of our experiments however, we were able to ignore all of the task specific information regarding the target position and focus on the sensory input alone. This allows us to be as general as possible including only the speed and position of each of the simulated motors as well as the $x,y,z$ velocity measurements and the roll and pitch measurements produced by the simulator as our state space, all sensor measurements that would be very commonly found on real robots.

\subsection{Task Planner}
In order to transfer the general knowledge of the self contained in the self-model to a more specific knowledge of tasks we use Proximal Policy Optimization (PPO) \cite{ppo} to train a policy for each task. PPO has been used in the past to learn robotic control policies in simulation and in the real world alike. We use PPO to learn a policy for controlling the robot by just using the self-model trained on the robot's movements. Our PPO agent has a policy network consisting of 2 fully connected layers of 64 hidden units each. This network while not particularly complex was sufficient to learn all our intended policies on real data and as such is sufficient for all our experiments.

In the normal reinforcement learning paradigm, a reinforcement learning agent is given a state and learns an appropriate action in order to maximize its reward over an episode. Normally, this state comes from the "real world" sometimes data from a real robot, but more often data from a hand-coded simulator. In this work, we instead give the data output from the self-model. 

In this paper we ran 2 experiments, both of which had the exact same setup except for the calculation of the reward function. In every experiment the first thing that is done is the simulated robot is "reset" to an initial state and that first observation is returned. This observation $S_0$ is given to the self-model as a seed observation and it is given to the agent as well. The agent then produces its first action $A_0$. This action is passed to the self-model who will then output the predicted next state $\hat{S_1}$. $\hat{S_1}$ is then passed to the agent who will produce $A_1$. This process will continue until the episode length is reached. 

For the 2 different experiments we changed the reward function in order to cause the agent to learn a new task. Unlike the state, we instead used the true reward function as the learning of reward functions is outside the scope of this research. The first task we learned was forward locomotion. In order to learn this task we use the $x$ velocity at each timestep as the reward so that the episode's final reward is the total distance traveled resulting in agents who move forward efficiently. The second task we learned was jumping, or vertical locomotion. For this task, we set our reward equal to the velocity in the $z$ direction with the added effect that if the agent passed a defined $z$ position then the episode would terminate. This was done for smoothness in the agent's learning. 

\begin{figure}[h]
	\begin{center}
	\includegraphics[scale=0.35]{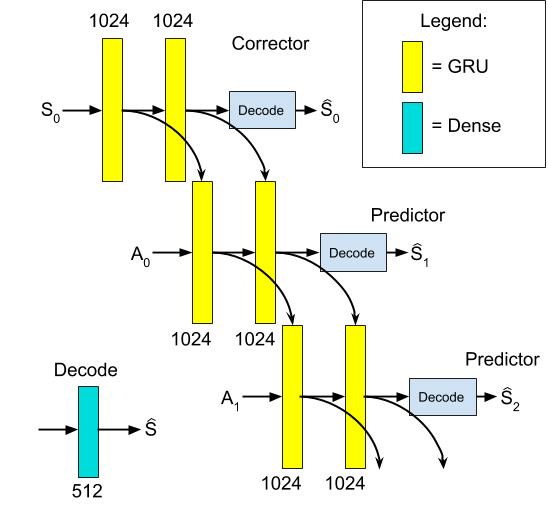}
	\end{center}
	\caption{The self-model architecture. To start the model is initialized by $S_0$. The next states $\hat{S}_{t+1}$ can be achieved by taking the model at state $t$ and passing as input action $A_t$. The model uses seperate recurrent networks for the Corrector and Predictor but shares the Decoder between the two.}
	\label{arch}
\end{figure}
\section{Model Training}
\subsection{Architecture}
Our self-model uses an architecture very similar to the one found in \cite{preco}. We use three seperate networks called the predictor, the corrector, and the decoder. The predictor and corrector networks are both recurrent networks of the same dimeisions whereas the decoder can be any non-recurrent network. In our case, we chose to use Gated Recurrent Units (GRUs) \cite{gru} as our recurrent networks and a traditional feed-forward network as our decoder the details of the networks can be seen in Figure \ref{arch}. The process of producing predicted state $\hat{S}$ makes use of all of these 3 networks. In order to produce the initial predicted state $\hat{S}_1$ we pass as inputs to the network $S_0$ as well as $A_1$. The corrector network is given as input $S_0$ and a null hidden state. The corrector produces a hidden state $H_0$ has its outputs piped to the decoder which outputs a reconstruction of the same state $\hat{S}_0$. The predictor then takes as input the action $A_1$ and uses the hidden state of the corrector $H_0$ as the hidden state of the RNN. This predictor then produces its own hidden state $H_1$ and has its output passed through the same decoder as in the previous step to produce the predicted next state $\hat{S}_1$. If no additional state information is received this process can be continued indefinitely by passing the next action $A_{t}$ and the prior hidden state $H_{t-1}$ to the predictor again to produce the predicted state $\hat{S}_t$. Additionally, if any new ground truth data is observed the network can again update through the corrector.

This architecture is able to leverage the states and the actions in a way that doesn't require an input state for each timestep. The recurrent units' hidden states also serve as a embedding space for a more abstract representation of state than the pure observation. Because the hidden state contains information from many past states it can have second order information that could only be understood by looking at many state action pairs. It is this information that allows the model to be effective for state prediction overtime. 


\subsection{Training Procedure}
We used the same self-model for every experiment. In order to train the self-model we first had to generate data. In an effort to demonstrate the transferability of the self-model, we generated the data in the most naive way possible, completely random actions. We sampled 100,000 actions from a uniform distribution and saved the corresponding $(S_t,A_t,S_{t+1})$ tuple. This data was used to train our model and we used another 5000 tuples as our validation set. In order to do this training, we had 3 mean square error (MSE) losses that were minimized. The first loss is a reconstruction loss between the predicted $\hat{S}_t$ and the real $S_t$. The second loss was the MSE between the predicted $\hat{S}_{t+1}$ and the true $S_{t+1}$ We also collected this data and extracted all sequences extending $n$ steps out from each state. For all of our experiments we set $n=100$. Using these sequences we calculate another loss using the MSE of the predicted versus true states for all $t+i$ in the sequence. The loss is then calculated as the average of the calculated MSE's.

\begin{align*}
L_{recon} &= mean((\hat{S}_{t}-S_{t})^2)\\
L_{single} &= mean((\hat{S}_{t+1}-S_{t+1})^2)\\
L_{seq} &= \frac{1}{n}\sum_{i=1}^{n}{mean((\hat{S}_{t+i}-S_{t+i})^2)}\\
\end{align*}

Our backpropigation then follows after having computed each of these losses. This system of 3 losses over the single step, sequence, and reconstruction is similar to the one used in \cite{preco}.

\begin{figure}
	\begin{center}
	\begin{tabular}{ccc}
		\includegraphics[scale=0.47]{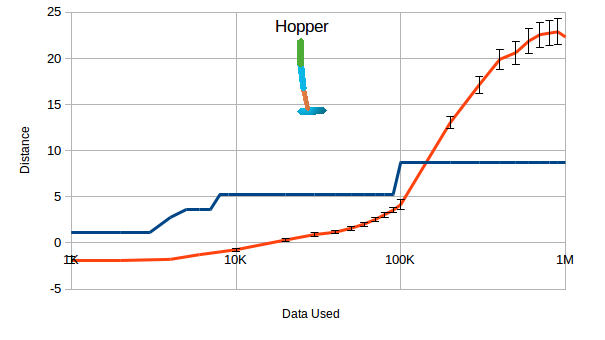}\\
		\includegraphics[scale=0.47]{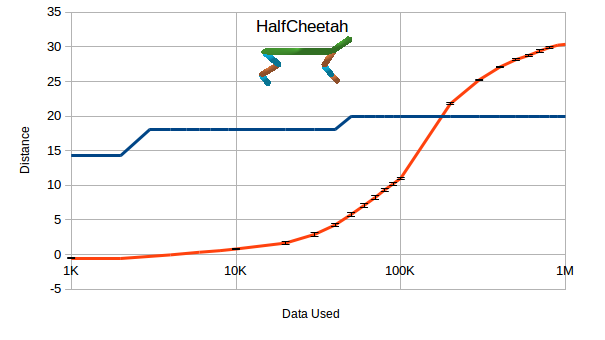}\\
		\includegraphics[scale=0.47]{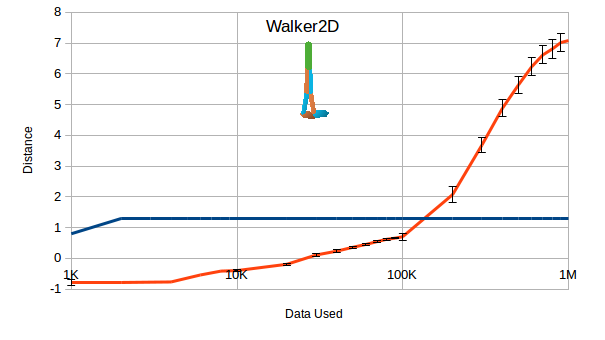}\\
		\includegraphics[scale=0.47]{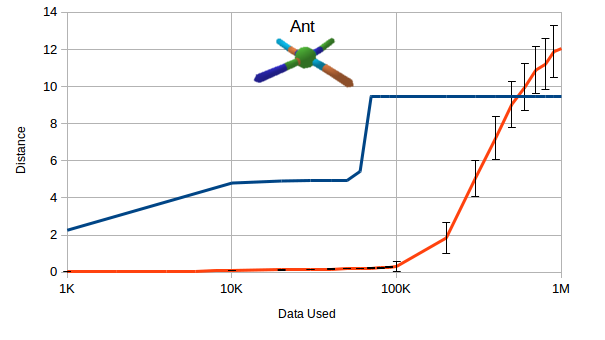}\\
		\includegraphics[scale=0.47]{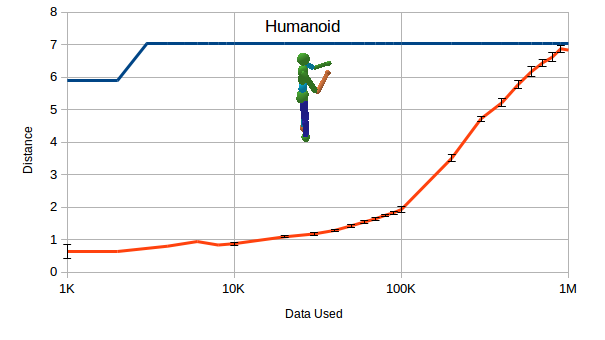}
	\end{tabular}
	\includegraphics[scale=0.66]{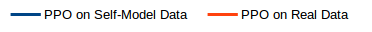}
	\end{center}
	\caption{Graphs of the distance traveled as a function of the data used. These graphs show the self-model trained up to 100,000 state action tuples and the PPO model trained up to 1,000,000}
\end{figure}
\section{Results}

\subsection{Learning to Walk}

We tested our algorithm using the same architecture and model configuration on a variety of different robot platforms found in the PyBullet API and demonstrated a significant improvement in data efficiency on every platform tested. For every experiment we trained the self-model with up to 100,000 real state action tuples whereas we trained the PPO agents to up to 1,000,000 state action tuples. When using comparable data the PPO agent trained on self-model artifical data always outperformed the PPO agent trained on real data. This suggests that the ability of the self-model to generalize on small amounts of data is strong and we were able to leverage this to allow the PPO agent to use more artificial data thus giving it more training cycles with the same amount of seen real state-action tuples.

The data efficiency gains were most perceivable when compared on relatively small amounts of data. When using PPO on a trained self-model we were able to achieve data reduction factors of up to 500 times. This is to say we were able to achieve the same result with one 500th the amount of data as PPO required when using real data, and were able to train an agent on the Humanoid environment on a self-model trained on only 3000 state-action tuples to perform as well as a PPO agent trained on 1,000,000 timesteps. Interestingly, we noticed a trend across our varied environments. We noticed that as the complexity of the system increased the data reduction factor also increased in a linear relationship with an $R^2$ of 0.99. This relationship is further strengthened by the fact that the HalfCheetah and Walker2D environments, despite having very different morphologies, have nearly identical data reduction factors. The existence of this phenomenon suggests that self-models would have a tremendous gain in the efficiency of training high complexity systems and could be a key piece in the expansion of reinforcement learning to more complicated domains.

\subsection{Comparison to State of the Art}
While we cannot do a direct comparison to other state of the art methods due to our modified reward function we can make comparisons in terms of data efficiency. When compared with the state of the art method TRPO \cite{trpo} we see that our method shows significant improvements using low and very low amounts of data. When we approach the higher ranges of data TRPO begins to outperform our algorithm and in some of the low dimensionality cases we only perform slightly better than PPO. This serves to emphasize that our algorithm is most effective when approaching the most challenging cases to traditional reinforcement learning, namely the low data and high dimensionality problems and in those cases our algorithm performs well above the state of the art.  

\begin{figure}
\hspace*{0.1in}
\begin{center}
	\begin{tabular}{|c|c|c|c|}
	\hline
	Experiment & Datapoints & TRPO & Self-Model\\
	\hline
	Hopper & 1K & 0 & \textbf{40}\\
	HalfCheetah & 1K & 18 & \textbf{118}\\
	Ant & 1K & 13 & \textbf{212}\\
	Humanoid & 1K & 0 & \textbf{526}\\
	\hline
	Hopper & 10K & 0 & \textbf{11.1} \\
	HalfCheetah & 10K & 10 & \textbf{14.7} \\
	Ant & 10K & 10 & \textbf{21.5}\\
	Humanoid & 10K & 30 & \textbf{33.9}\\
	\hline
	Hopper & 100K & \textbf{100} & 1.5 \\
	HalfCheetah & 100K & \textbf{10} & 1.67\\
	Ant & 100K & \textbf{100} & 5.5\\
	Humanoid & 100K & 9 & \textbf{10}\\
	\hline
	\end{tabular}
	\caption{A comparison of the data efficiency of our method and TRPO, the current state of the art. The TRPO and Self-Model columns correspond to what multiple of the data used to train the World Model vanilla PPO requires to achieve the same results.}
	\label{trpo_table}
\end{center}
\end{figure}

\subsection{Learned Gait}
\begin{figure}[h]
\hspace*{0.3in}
	\begin{center}
	\includegraphics[scale=0.32]{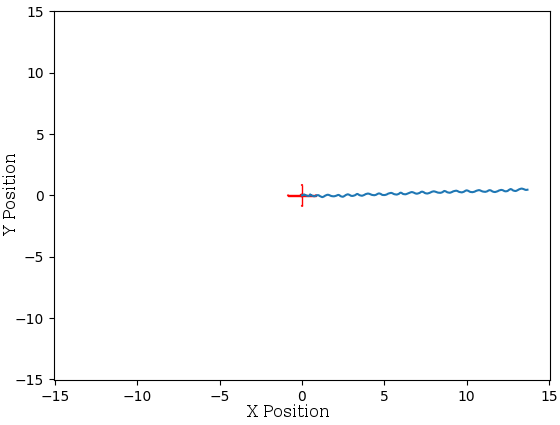}
	\end{center}
	\caption{Distance traveled in the Ant environment by the self-model trained agent (blue) compared to the body of the agent (red).}
	\label{walk_dist}
\end{figure} 
\begin{figure}[h]
	\begin{center}
	\includegraphics[scale=0.6]{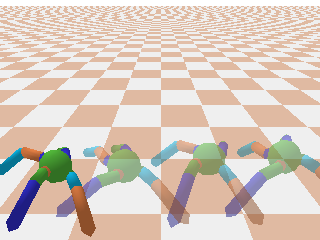}
	\end{center}
	\caption{Walking policy learned by the agent trained on self-model.}
	\label{walking_policy}
\end{figure}

Despite training entirely in self-model, the agents learn an effective gait and one that can translate well into the real world. Figure \ref{walk_dist} shows that the path traveled by the Ant robot, shown in blue, far exceeds the size of the robot shown as the red cross. Similarly, the path here is almost entirely straight with only minor deviations in a wave-like pattern. These deviations are to be expected as the position reported corresponds to the center of the robot, and while moving it has a tendency to sway back and forth. 
This gait itself is very similar to those trained using the real simulator. The learned gait also progressed in a fairly natural way, and one similar to those gaits learned on real simulators. At 100,000 steps the policy had begun to learn how to catch itself from falling in an effective manner, and that moving one of its legs is a useful strategy. It first moves its one leg back and forth and can be seen in dark blue. At 500,000 steps the policy has also noticably improved. The robot has learned to make use of all 4 of its legs to land and immediately start to take its first steps. However, after catching itself from falling the robot resumes the inefficient policy of moving one leg back and forth. As the agent approaches 1,000,000 steps the gait becomes more general. Instead of taking one step in the beginning of the episode the agent has learned how to take a step and from that position take another generalizing into a complete walking gait. This timeframe to achieve success mirrors the process that the agent learning on the real environment takes suggesting that the self-model is a close approximation to the real environment.

\subsection{Learning to Jump}
One of the most important benefits of the self-model based approach to training agents is its ability to immediately shift to learning a new task without the need to collect any new state action tuples. Our trained agent showed that it was able to learn a jumping policy learning using the self-model in open loop. The agent learns a policy that is noticably different from the walking policies. The agent first catches itself with its legs much straighter than before and then lifts the legs off of the ground to produce some momentum such that it can eventually propel itself upward and then fall gracefully back down to the ground so that it can do it again.

\begin{figure}[h]
	\begin{center}
	\includegraphics[scale=0.6]{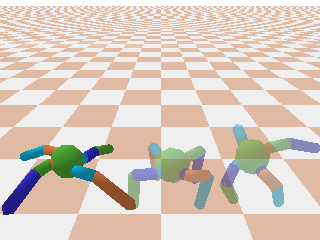}
	\end{center}
	\caption{Jumping policy learned by the agent trained on self-model.}
	\label{jumping_policy}
\end{figure}
\begin{figure}[h]
	\begin{center}
	\includegraphics[scale=0.35]{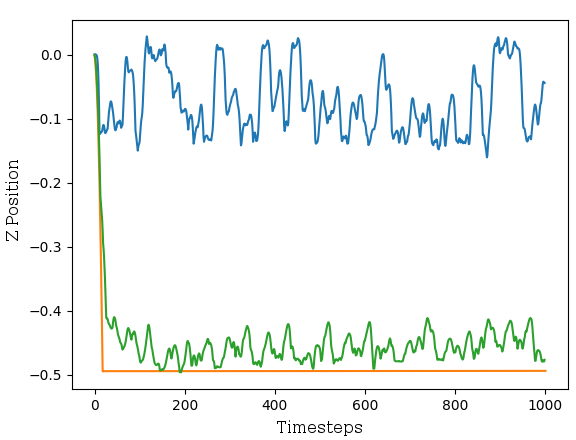}
	\end{center}
	\caption{Z motion over time comparing a jumping policy (blue) a walking policy (green) and an untrained policy (orange).}
	\label{z_chart}
\end{figure}

The success of the jumping policy is even more notable when the motion in the vertical ($z$) direction is plotted over the duration of the episode. In Figure \ref{z_chart} we show the $z$ motion for a the jumping policy trained on the self-model, the running policy trained on the self-model and an untrained policy as a baseline. The blue chart (the policy trained to jump) shows not only a significantly higher average $z$ position but a significantly higher variance than both of the other policies. The higher average $z$ position shows it has successfully achieved its goal outlined by its reward function which rewarded it for moving higher. The maximum heights shown also suggests that the agent has successfully learned to jump. The agent in this situation would not be able to go higher than a height of $0$ with its feet on the ground. The successes shown in this experiment further outline the power of the self-model which was able to go seamlessly from learning to walk to learning a completely new and seperate task.

\section{Conclusion}
Self-models are very powerful tools for learning. If used properly, a self-model can, through using significantly fewer state action tuples, learn a variety of tasks when coupled with a reinforcement learning agent. This self-model can leverage any data from any task even completely random motion. As such gathering data for training the self-model does not necessitate the agent trying to learn any particular task. If the self-model is properly trained on the ground truth data a reinforcement learning agent can train itself using the self-model as an environment. If a reward function can too be simulated for the reinforcement learning agent, then such an agent could give the appearance of zero-shot learning by training to do any number of tasks without the need to collect any additional data.

The self-models presented here could also benefit from more research on combining environment learning techniques like our self-model with reinforcement learning or task planning as work like \cite{world_models} \cite{recurrent_env_sims} and \cite{ref_i2a} have done in the past. Furthermore, incorporating reward prediction within the self-model and allowing all reinforcement learning to be trained completely end-to-end on a self-model could bolster or circumvent entirely the need for these techniques. Through leveraging these techniques along with our demonstrated powerful self-modeling techniques we hope that we can make reinforcement learning and lifelong learning a reality.\\
\subsubsection*{Acknowledgments}
Supported by DARPA MTO grant HR0011-18-2-0020
\newpage

\bibliographystyle{IEEEtran}
\bibliography{references}

\begin{thebibliography}{10}
\providecommand{\url}[1]{#1}
\csname url@rmstyle\endcsname
\providecommand{\newblock}{\relax}
\providecommand{\bibinfo}[2]{#2}
\providecommand\BIBentrySTDinterwordspacing{\spaceskip=0pt\relax}
\providecommand\BIBentryALTinterwordstretchfactor{4}
\providecommand\BIBentryALTinterwordspacing{\spaceskip=\fontdimen2\font plus
\BIBentryALTinterwordstretchfactor\fontdimen3\font minus
  \fontdimen4\font\relax}
\providecommand\BIBforeignlanguage[2]{{%
\expandafter\ifx\csname l@#1\endcsname\relax
\typeout{** WARNING: IEEEtran.bst: No hyphenation pattern has been}%
\typeout{** loaded for the language `#1'. Using the pattern for}%
\typeout{** the default language instead.}%
\else
\language=\csname l@#1\endcsname
\fi
#2}}

\bibitem{gazebo}
N.~Koenig and A.~Howard, ``Design and use paradigms for gazebo, an open-source
  multi-robot simulator,'' in \emph{2004 IEEE/RSJ International Conference on
  Intelligent Robots and Systems (IROS)(IEEE Cat. No. 04CH37566)},
  vol.~3.\hskip 1em plus 0.5em minus 0.4em\relax IEEE, 2004, pp. 2149--2154.

\bibitem{josh_science}
J.~Bongard, V.~Zykov, and H.~Lipson, ``Resilient machines through continuous
  self-modeling,'' \emph{Science}, vol. 314, no. 5802, pp. 1118--1121, 2006.

\bibitem{rob_science}
\BIBentryALTinterwordspacing
R.~Kwiatkowski and H.~Lipson, ``Task-agnostic self-modeling machines,''
  \emph{Science Robotics}, vol.~4, no.~26, 2019. [Online]. Available:
  \url{https://robotics.sciencemag.org/content/4/26/eaau9354}
\BIBentrySTDinterwordspacing

\bibitem{preco}
\BIBentryALTinterwordspacing
B.~Amos, L.~Dinh, S.~Cabi, T.~Roth{\"{o}}rl, S.~G. Colmenarejo, A.~Muldal,
  T.~Erez, Y.~Tassa, N.~de~Freitas, and M.~Denil, ``Learning awareness
  models,'' \emph{CoRR}, vol. abs/1804.06318, 2018. [Online]. Available:
  \url{http://arxiv.org/abs/1804.06318}
\BIBentrySTDinterwordspacing

\bibitem{world_models}
\BIBentryALTinterwordspacing
D.~Ha and J.~Schmidhuber, ``World models,'' \emph{CoRR}, vol. abs/1803.10122,
  2018. [Online]. Available: \url{http://arxiv.org/abs/1803.10122}
\BIBentrySTDinterwordspacing

\bibitem{recurrent_env_sims}
\BIBentryALTinterwordspacing
S.~Chiappa, S.~Racani{\`{e}}re, D.~Wierstra, and S.~Mohamed, ``Recurrent
  environment simulators,'' \emph{CoRR}, vol. abs/1704.02254, 2017. [Online].
  Available: \url{http://arxiv.org/abs/1704.02254}
\BIBentrySTDinterwordspacing

\bibitem{ref_i2a}
\BIBentryALTinterwordspacing
T.~Weber, S.~Racani{\`{e}}re, D.~P. Reichert, L.~Buesing, A.~Guez, D.~J.
  Rezende, A.~P. Badia, O.~Vinyals, N.~Heess, Y.~Li, R.~Pascanu, P.~Battaglia,
  D.~Silver, and D.~Wierstra, ``Imagination-augmented agents for deep
  reinforcement learning,'' \emph{CoRR}, vol. abs/1707.06203, 2017. [Online].
  Available: \url{http://arxiv.org/abs/1707.06203}
\BIBentrySTDinterwordspacing

\bibitem{policy_opt_exp}
\BIBentryALTinterwordspacing
F.~Pan, Q.~Cai, A.~Zeng, C.~Pan, Q.~Da, H.~He, Q.~He, and P.~Tang, ``Policy
  optimization with model-based explorations,'' \emph{CoRR}, vol.
  abs/1811.07350, 2018. [Online]. Available:
  \url{http://arxiv.org/abs/1811.07350}
\BIBentrySTDinterwordspacing

\bibitem{model_based_atari}
\BIBentryALTinterwordspacing
L.~Kaiser, M.~Babaeizadeh, P.~Milos, B.~Osinski, R.~H. Campbell, K.~Czechowski,
  D.~Erhan, C.~Finn, P.~Kozakowski, S.~Levine, R.~Sepassi, G.~Tucker, and
  H.~Michalewski, ``Model-based reinforcement learning for atari,''
  \emph{CoRR}, vol. abs/1903.00374, 2019. [Online]. Available:
  \url{http://arxiv.org/abs/1903.00374}
\BIBentrySTDinterwordspacing

\bibitem{trpo}
\BIBentryALTinterwordspacing
T.~Kurutach, I.~Clavera, Y.~Duan, A.~Tamar, and P.~Abbeel, ``Model-ensemble
  trust-region policy optimization,'' \emph{CoRR}, vol. abs/1802.10592, 2018.
  [Online]. Available: \url{http://arxiv.org/abs/1802.10592}
\BIBentrySTDinterwordspacing

\bibitem{uncert_imagination}
\BIBentryALTinterwordspacing
G.~Kalweit and J.~Boedecker, ``Uncertainty-driven imagination for continuous
  deep reinforcement learning,'' in \emph{Proceedings of the 1st Annual
  Conference on Robot Learning}, ser. Proceedings of Machine Learning Research,
  S.~Levine, V.~Vanhoucke, and K.~Goldberg, Eds., vol.~78.\hskip 1em plus 0.5em
  minus 0.4em\relax PMLR, 13--15 Nov 2017, pp. 195--206. [Online]. Available:
  \url{http://proceedings.mlr.press/v78/kalweit17a.html}
\BIBentrySTDinterwordspacing

\bibitem{pybullet}
E.~Coumans and Y.~Bai, ``Pybullet, a python module for physics simulation for
  games, robotics and machine learning,'' \url{http://pybullet.org}, 2019.

\bibitem{MAML}
\BIBentryALTinterwordspacing
C.~Finn, P.~Abbeel, and S.~Levine, ``Model-agnostic meta-learning for fast
  adaptation of deep networks,'' \emph{CoRR}, vol. abs/1703.03400, 2017.
  [Online]. Available: \url{http://arxiv.org/abs/1703.03400}
\BIBentrySTDinterwordspacing

\bibitem{ant_obstacles}
\BIBentryALTinterwordspacing
N.~Heess, D.~TB, S.~Sriram, J.~Lemmon, J.~Merel, G.~Wayne, Y.~Tassa, T.~Erez,
  Z.~Wang, S.~M.~A. Eslami, M.~A. Riedmiller, and D.~Silver, ``Emergence of
  locomotion behaviours in rich environments,'' \emph{CoRR}, vol.
  abs/1707.02286, 2017. [Online]. Available:
  \url{http://arxiv.org/abs/1707.02286}
\BIBentrySTDinterwordspacing

\bibitem{ppo}
\BIBentryALTinterwordspacing
J.~Schulman, F.~Wolski, P.~Dhariwal, A.~Radford, and O.~Klimov, ``Proximal
  policy optimization algorithms,'' \emph{CoRR}, vol. abs/1707.06347, 2017.
  [Online]. Available: \url{http://arxiv.org/abs/1707.06347}
\BIBentrySTDinterwordspacing

\bibitem{gru}
\BIBentryALTinterwordspacing
K.~Cho, B.~van Merrienboer, {\c{C}}.~G{\"{u}}l{\c{c}}ehre, F.~Bougares,
  H.~Schwenk, and Y.~Bengio, ``Learning phrase representations using {RNN}
  encoder-decoder for statistical machine translation,'' \emph{CoRR}, vol.
  abs/1406.1078, 2014. [Online]. Available:
  \url{http://arxiv.org/abs/1406.1078}
\BIBentrySTDinterwordspacing

\end{thebibliography}

\end{document}